\documentclass[11pt]{article}

\usepackage{fullpage}
\usepackage{doi}

\usepackage{amsmath,amsthm}
\usepackage{amsfonts,amssymb}

\usepackage{microtype}

\usepackage{latexsym}

\usepackage{graphicx}

\usepackage{subfig}\DeclareCaptionType{copyrightbox}
\usepackage{xcolor}

\graphicspath{{./figs/}}

\usepackage{wrapfig,threeparttable}

\usepackage{rotating}

\usepackage{verbatim}
\usepackage{multirow}

\usepackage[ruled,linesnumbered]{algorithm2e}

\newtheorem{defi}{Definition}

\newcommand{\En}{\Phi}

\begin{document}

\title{\LARGE\bf
Particles Prefer Walking Along the Axes:
Experimental Insights into the Behavior of a Particle Swarm}

\author{Manuel Schmitt \qquad Rolf Wanka\\[2mm]
Department of Computer Science\\
University of Erlangen-Nuremberg, Germany\\
{\{\tt manuel.schmitt, rolf.wanka\}@cs.fau.de}
}

\date{ }

\maketitle

\begin{abstract}
Particle swarm optimization (PSO)
is a widely used 
nature-inspired meta-heuristic for solving
continuous optimization problems.
However, when running the PSO algorithm, one encounters the 
phenomenon of so-called
stagnation, that means in our context, the whole swarm starts to converge
to a solution that is not (even a local) optimum.
The goal of this work is to point out
possible reasons why the swarm stagnates at these
non-optimal points.
To achieve our results, we use the newly
defined \emph{potential} of a swarm.
The total potential has a portion for every dimension of
the search space, and it drops when
the swarm approaches the point of convergence.
As it turns out experimentally, the swarm is very likely to come
sometimes into ``unbalanced'' states, i.\,e., almost all
potential belongs to one axis. Therefore, the swarm becomes
blind for improvements still possible in any other direction.
Finally, we show how in the light of the potential and these
observations, a slightly adapted PSO rebalances
the potential and therefore increases the quality of the
solution. Note that this is an extended version of \cite{SWb:13}.
\end{abstract}


\section{Introduction}
\label{sec:intro}
%
%

\noindent\textbf{\sf Background.}
Particle swarm optimization (PSO), introduced by Kennedy and 
Eberhart~\cite{ken_eb_1995,eb_ken_1995}, is a very popular 
meta-heuristic for solving continuous optimization problems.
It is inspired by the social interaction of individuals living
together in groups and supporting and cooperating with each other. 
Fields of very successful application are, among many others,
in Biomedical Image Processing~\cite{WSZZE:04},
Geosciences~\cite{OD:10},
Mechanical Engineering~\cite{GWHK:09},
and Materials Science~\cite{RPPN:09}, to name just a few, where the
continuous objective function on a multi-dimensional domain
is not given in a closed form, but by a ``black box.''
The popularity of the PSO framework in these scientific communities
is due to the fact that it on the one hand can be realized and,
if necessary, adapted to further needs easily,
but on the other hand shows in experiments good performance results with respect
to the quality of the obtained solution and the speed needed to obtain it.
By adapting its parameters, users may in real-world applications
easily and successfully control the swarm's
behavior with respect to
``exploration'' (``searching where no one has searched before'')
and
``exploitation'' (``searching around a good position'').
A thorough discussion of PSO can be found in~\cite{swarmhandbook:11}.

To be precise, let an objective function $f:\mathbb{R}^D\rightarrow \mathbb{R}$
on a $D$-dimensional domain be given
that (w.\,l.\,o.\,g.) has to be minimized.
A population of {\it particles}, each consisting of a position (the candidate for
a solution), a velocity
and a local attractor, moves through the search space $\mathbb{R}^D$. The
local attractor of a particle is the best position with respect to $f$
this particle has encountered so far.
The population in motion is the \emph{swarm}.
In contrast to other evolutionary algorithms,
the individuals of the swarm cooperate 
by sharing information about the search space
via the global attractor, which is the best position any particle has found
so far. The particles move in time-discrete iterations. 
The movement of a particle is governed 
by both its velocity and the two attractors and by some additional fixed parameters
(for details, 
see Sec.~\ref{sec:definitions}).

PSO is widely used in real-world applications.
It is usually desired that the swarm converges to
a single point in the search space, and that this
point is at least a local optimum.
It is well investigated how the fixed parameters mentioned above
should be chosen to let the swarm
converge at all~\cite{T:03,JLY:07a}\footnote{In the companion paper \cite{SWa:13} to
the paper at hand, the quality of the best point found by the
swarm (the global attractor) is formally analyzed.}.
However, experiments sometimes show
the phenomenon of stagnation, meaning here that
the particle swarm sometimes
converges to a single search point and gets stuck, although this
point is not even a local optimum.
In~\cite{LW:11}, Lehre/Witt 
have for a certain
situation formally proven that the probability for a swarm to converge to a non-optimal
point is positive.
Several approaches to deal with this stagnation phenomenon have been developed.
Clerc~\cite{C:06} 
examines the distribution
of the particles during the stagnation phase,
derives properties of these distributions,
and provides several possibilities to adapt the algorithm for the
case that the number of iterations without an improvement
reaches a certain threshold.
%
Van den Bergh/Engelbrecht~\cite{BE:02} substantially modify
the movement equations, enabling the particles to count the number of times they
improve the global attractor and use this information.
Empirical evidence for the capability of
their method to find local optima on common benchmarks is given.
Closest to our work, in~\cite{LW:11} the movement equations are modified
by adding in \emph{every} iteration a small random perturbation to the velocity.


\medskip

\noindent\textbf{\sf New results.}
In this paper, we focus on the swarm's behavior right before the convergence starts,
in order to find out about possible causes that let the swarm converge,
i.\,e., why the global attractor starts to stagnate. 
Although one would like the output of the PSO algorithm to be at least a local optimum, we
point out two possible reasons for a swarm to converge far away from any local optimum. In 
order to state the causes for this premature convergence, we define a \emph{potential} that
reflects the capability of the swarm to move. That means the swarm converges iff
the total potential approaches $\vec 0$.
The swarm's total potential has a portion for every dimension of the
$D$-dimensional search space.
The experiments carried out suggest that unwanted stagnation 
can indeed be explained in terms of the potential.

The first possible cause of a non-optimal limit of convergence
is that, even though the global attractor is updated frequently, the potential drops,
so the swarm has not sufficient momentum to find significantly improved points.
In Sec.~\ref{sec:1Dim}, we present experiments that demonstrate that
this phenomenon can in fact be observed for some parameter selections.
Fortunately, it also turns out that common parameter choices and a reasonable swarm
size already avoid this problem.

The second and more important reason is that the potential tends to becoming imbalanced
among the dimensions, so dimensions in which
only small improvement is possible may neverheless have the by far highest potential.
That means that every step in
such a dimension results in a worsening strong enough to void possible improvements in other
dimensions. So, the global attractor of the swarm stagnates and the swarm starts to converge.
We demonstrate that indeed the swarm tends to reach a state where the potentials are unbalanced,
i.\,e., one dimension gets the by far highest portion of the total
potential while all other portions are about equal. Then,
we present experimental evidence showing that this phenomenon makes the particles converge 
at non-optimal search points.
So the experiments suggest that 
first the swarm starts to prefer a promising direction that is parallel to
one of the axes and increases the potential in this dimension far above the
potentials in the other dimensions. As soon as the chosen direction does no
longer yield improvements, its potential stays much larger than in the 
remaining dimensions where improvements would still be possible. From that point on,
improvements become rare and the swarm starts to stagnate, although no local optimum
is found yet.

Since the cause of this premature convergence is an imbalance of the potentials, we show how
a small, simple and easy to implement modification of the algorithm enables it to handle such situation. 
Namely, if the potential is sufficiently small, we let the particles make pure random steps, which do not
favor any direction. We conclude with showing that the modification does not totally overwrite the
PSO algorithm by replacing it by some greedy random search procedure. Instead, our experiments show that
the modification is only applied in case of (premature) convergence. As long as there is still
room for improvements left in the search space, the unmodified behavior prevails. 



 \section{Definitions}
\label{sec:definitions}
First we present the underlying model of the PSO algorithm.
The model describes the positions of the particles, the velocities and the global and local
attractors as real-valued stochastic processes. Furthermore, we define in Def.~\ref{Energy}
the potential of a swarm that will be a measure for its movement.
A swarm with high potential is more likely to reach search points far away from the current
global attractor, while a swarm with potential approaching $0$ is converging.

\begin{defi}[Classical PSO Algorithm]\label{cPSO}
A \emph{swarm} $\cal S$ of $N$ particles moves through the $D$-dimensional search space
$\mathbb{R}^D$. Each particle $n$ consists of the following components:
\begin{itemize}
\item \emph{position} $X^n\in\mathbb{R}^D$, describing the current location of the particle in the search space,
\item \emph{velocity} $V^n\in\mathbb{R}^D$, describing the vector of the particle's current velocity, 
\item \emph{local attractor} $L^n\in\mathbb{R}^D$, describing the best point particle $n$ has visited so far.
\end{itemize}
Additionally, the swarm shares information via the \emph{global attractor} $G\in\mathbb{R}^D$, describing the best 
point \emph{any} particle has visited so far.
%
After some arbitrary initialization of the $X^n$ and $V^n$ (usually, one assumes them
to be initialized u.\,a.\,r. over some domain), the actual movement of the swarm is governed 
by the procedure described in Algorithm~\ref{alg:clPSO} where $f:\mathbb{R}^D\rightarrow \mathbb{R}$
denotes the objective function.
\begin{algorithm}[tbh]
\caption{\label{alg:clPSO}classical PSO}

\SetKwInOut{Input}{input}
\SetKwInOut{Output}{output}

\Output{$G\in\mathbb{R}^D$}
\Repeat{termination criterion met}{
  \For{$n=1 \to N$}{
    \For{$d=1 \to D$}{
      $V_d^n := \chi\cdot V_d^n + c_1\cdot \operatorname{rand}() \cdot (L_d^n-X_d^n)\newline
	 \hphantom{V_d^n := \chi\cdot V_d^n} + c_2\cdot \operatorname{rand}() \cdot (G_d-X_d^n)$\label{velup}\;
      $X_d^n := X_d^n+V_d^n$\;
    }
    \If{$f(X^n)\le f(L^n)$}{
      $L^n := X^n$\;
    }
    \If{$f(X^n)\le f(G)$}{
      $G := X^n$\;
    }
  }
}
\end{algorithm}

Here, $\chi$, $c_1$ and $c_2$ are some positive constants called the fixed parameters
of the swarm, 
and $\operatorname{rand}()$ returns values that are uniformly distributed over $[0,1]$ and all independent.

\end{defi}

Note that in case of a tie between the previous attractor and the new point $X^n$, we use the new value, i.\,e., whenever a 
particle finds a search point with value equal to the one of its local attractor, this point becomes
the new local attractor. If additionally the function value is equal to the one of the global attractor, 
this one is also updated.
Also note that the global attractor is updated as soon as
a new better solution has been found.

Now we want to define a potential for measuring how close the swarm is to convergence. A meaningful potential should of course involve the velocities of the particles. 
These considerations lead to the following definition:

\begin{defi}[Potential]\label{Energy}
For $d\in\{1,\ldots,D\}$, the potential of swarm $\cal S$ in dimension $d$ is
$\En_d$ with
$$
\En_d:=\sum_{n=1}^N\Big(|V_d^n|+|G_d-X_d^n|\Big)\enspace,
$$
the total potential of $\cal S$ is $\Phi=(\Phi_1,\ldots,\Phi_D)$.
\end{defi}

Note that we slightly deviate from the notion of the potential found in \cite{SWa:13} since the version in the 
definition above is simpler and sufficient for the present work. However, the underlying idea is the same
for both versions of the potential.

So the current total potential of a swarm
has a portion in every dimension.
Between two different dimensions, the potenial may differ much,
and ``moving'' potential from one dimension to another is not possible.
On the other hand, along the same dimension the particles
influence each other and can transfer potential from one to the other.
This is the reason why we do not define a potential of an individual particle.


\section{1-Dimensional PSO}
\label{sec:1Dim}

In this section, we examine the behavior of a $1$-dimensional PSO with respect to the potential.
If the swarm is close to a local optimum and there is no second local optimum within range, the
attractors converge and it is well-known that with appropriate choices for the parameters of
the PSO, convergence of the attractors implies convergence of the whole swarm. Such parameter
selection guidelines can be found, e.\,g., in \cite{JLY:07a}. 

On the other hand, if the swarm is far away from the next local optimum and the function is monotone
on an area that is large compared to the current potential of the swarm, the preferred behavior of the 
swarm is to increase the potential and move in the direction
that yields the improvement until a local optimum is surpassed and the monotonicity of the function changes.
In \cite{LW:11}, the authors show
that there are non-trivial choices of parameters for which the swarm converges even on a monotone function. In particular,
if $N=1$, 
every parameter choice either allows convergence to an arbitrary point in the search space, or it
generally prevents the one-particle-swarm from converging, even if the global attractor is already at the global
optimum.

\begin{figure}[thb]
\centering
\includegraphics[width=9.6cm
]{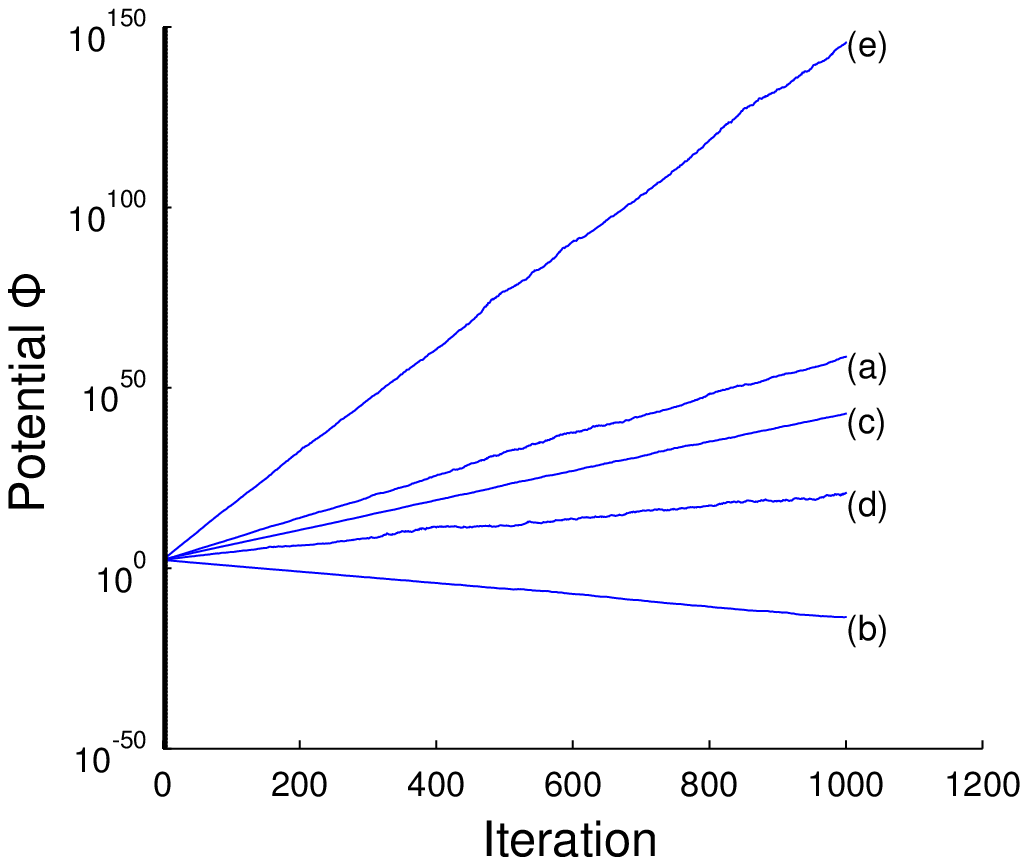}
\centerline{\parbox{0.5\textwidth}{\caption{\small\label{energie}\newline(a) $\chi = 0.729$, $c_1 = c_2 = 1.49$, $N=2$ \cite{CK:02} \newline
				(b) $\chi = 0.729$, $c_1 = 2.8\cdot \chi$, $c_2 = 1.3\cdot \chi$, $N=2$~\cite{Carlisle01}\newline
				(c) $\chi = 0.729$, $c_1 = 2.8\cdot \chi$, $c_2 = 1.3\cdot \chi$, $N=3$~\cite{Carlisle01}\newline
				(d) $\chi = 0.6$, $c_1 = c_2 = 1.7$, $N=2$ \cite{T:03}\newline
				(e) $\chi = 0.6$, $c_1 = c_2 = 1.7$, $N=3$ \cite{T:03}}}}
\end{figure}

We ran the particle swarm algorithm on a monotone function to measure the development of the potential over time.
For our experiment, we chose the $1$-dimensional function $f(x) = -x$ as objective function
wanting the swarm always ``running down the hill.''
Note that this choice is not a restriction,
since the particles compare points only qualitatively and the behavior is exactly the same
on any monotone decreasing function:
due to the rules for updating the attractors in lines 7 and 11, resp.,
of Algorithm~\ref{alg:clPSO},
the new attractors are the points with greater $x$-coordinate.
Therefore, we used only one function in our experiment. The parameters for the movement equations are common choices obtained from the literature. We let the particles make $1000$ iterations and stored the potential at every iteration. We made a total of $1000$ experiments for each set of parameters and calculated both average and standard deviation. The averages are stated in Fig.~\ref{energie}, the standard deviations are of the same order and therefore omitted. In cases (a), (c), (d) and (e), the particles have shown the expected behavior, namely an exponential increase of the potential.
So the swarm keeps running down the hill which is what we want it to do.

However, in case (b) where only two particles are involved,
we see the potential decreasing exponentially
because the number of particles is presumably too small.
In this case, the swarm will eventually stop, i.\,e., stagnate.
But we also see in case (c) that using one additional
particle and not changing the remaining parameters,
the swarms keeps its motion.

In all cases, for the small swarm size of $\ge 3$,
the common parameter choices avoid
the problem mentioned in~\cite{LW:11}.


\section{\textit{\Large\bfseries D}-Dimensional PSO}
\label{sec:DDim}

In the $D$-dimensional case, the situation is more complicated as now
the relations between distinct dimensions become important. 
A new problem arising is the following:
Assume that the whole swarm is close to a point $x\in\mathbb{R}^D$ such that
every change of the first coordinate leads to a significantly worse value of the objective function, while in the other dimensions there is still room for improvements. Furthermore let the swarm have high potential in the first and low potential in any other dimension. Then an improvement of the global attractor is still possible, but it is very unlikely and between two updates are many steps without an update. The reason is that any improvement in some of the dimensions $2, ..., D$ is voided by the much larger worsening in dimension $1$. It follows that the attractors stay constant for long times between two updates and so the swarm tends to converge and therefore looses potential. As long as the global attractor stays constant, the situation is symmetric in every dimension, so while converging, the imbalance is still maintained.

First, we want to examine if and how such imbalances arise. Assume that
the fitness function is (on some area) monotone in every dimension. One of our main observations is
that indeed in such a situation the swarm tends to pick one dimension
and favor it over all the others. As a consequence, the movement of the
swarm becomes more and more parallel to one of the axes.

\begin{figure}[htb]
\centering
\subfloat[\label{imb1}Fitness function $f$]
{\includegraphics[width=8cm,height=8cm]{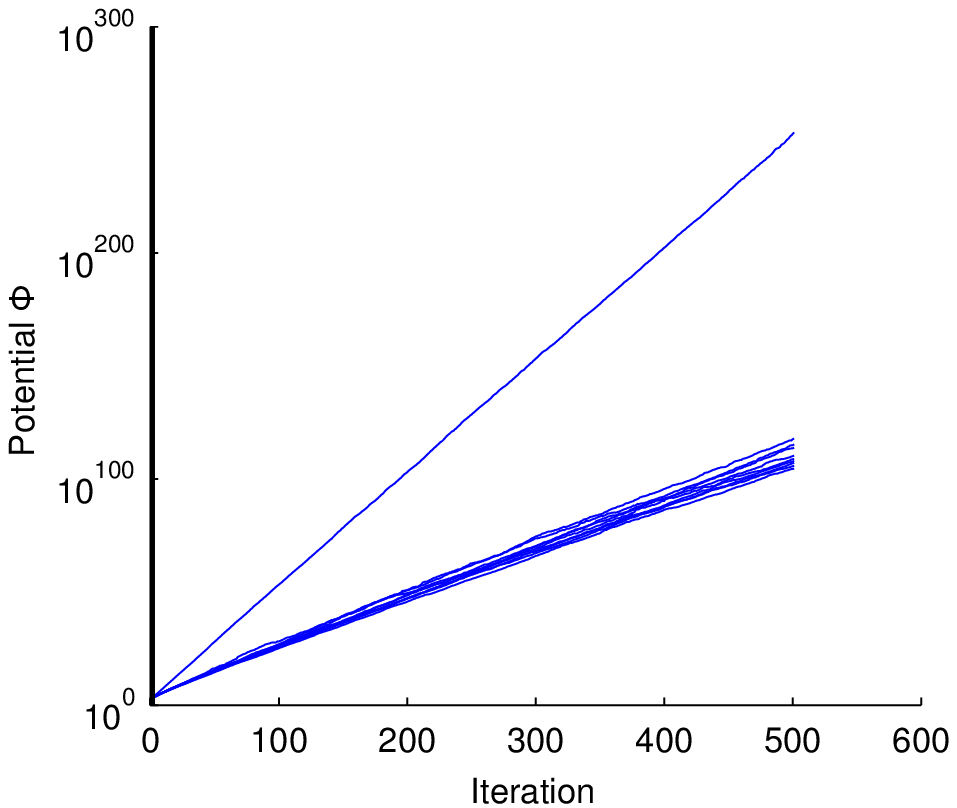}}
\subfloat[\label{imb2}Fitness function $g$]
{\includegraphics[width=8cm,height=8cm]{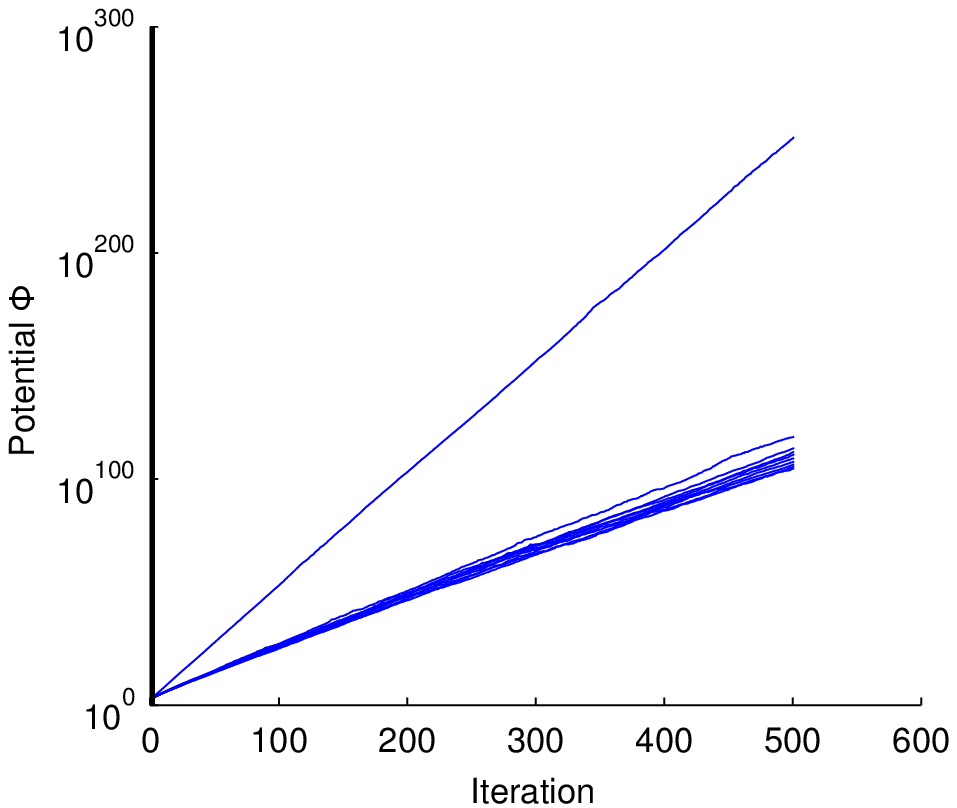}}
\caption{Growth of potential when processing (a) $f(\vec x) = -\sum_{i=1}^D x_i$, (b) $g(\vec x) = -\sum_{i=1}^D i\cdot x_i$.}
\label{imbalanced}
\end{figure}

We used the fitness-functions $f(\vec x) = -\sum_{i=1}^D x_i$ and $g(\vec x) = -\sum_{i=1}^D i\cdot x_i$
which are both monotonically decreasing in
every dimension and set $D$ to $10$. Initially, we distribute the particles randomly over $[-100; 100]^D$ and the
velocities over $[-50; 50]^D$ and let the swarm make $500$ iterations. The swarm size $N$ was $10$ and the parameters 
were set to $\chi = 0.729$, $c_1 = c_2 = 1.49$ as found in~\cite{CK:02}. After each iteration, we
calculated the potential for each dimension. We made $1000$ runs and for each run, the
dimensions were sorted according to the final value of $\En$, i.\,e., we switched the numbers of the dimensions such 
that after the last iteration dimension $1$ always had the highest potential, dimension $2$ the second highest and so on. 
We calculated the mean of the potentials over the $1000$ runs for each of the sorted dimensions. The results are stated in Fig. \ref{imbalanced}. One can see 
that the dimension with the greatest potential has for both functions a value far higher than the others, while the
other dimensions do not show such a significant difference between each other. In other words: Particles like to move parallel to an axis.

\begin{figure}[htb]
\centering
\includegraphics[width=7.2cm,height=5.6cm]{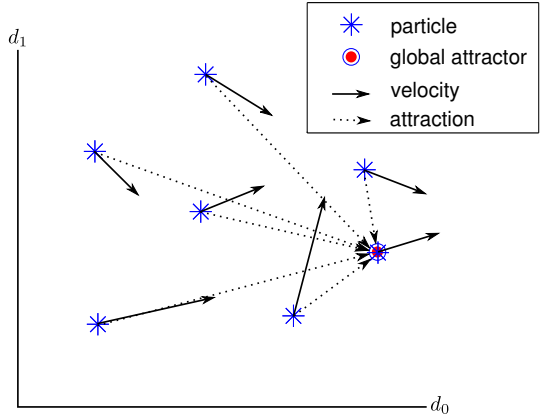}
\caption{Particles running in direction $d_0$. In dimension $d_0$, the differences between the coordinate of
the particle and the global attractor is on average higher than in dimension $d_1$. The velocities of dimension
$d_0$ point in the direction of the global attractor.}
\label{PotBild}
\end{figure}

An explanation for this behavior is the following: Assume that at some time, one dimension $d_0$ has
more potential than the others. Further assume that the advance is great enough such that for some
number of steps the particle with the largest value in dimension $d_0$ is the one that determines the
global attractor. In a companion paper, we call a swarm in this situation ``running''. Since randomness 
is involved and this situation has a positive probability to occur, it will actually occur after
sufficiently many iterations. Then, each update
of the global attractor increases the potential in $d_0$ considerably because it increases the distance
of every single particle to the global attractor except for the one particle that updated it. In any other
dimension $d\neq d_0$, the situation is different. Here, the decision which particle updates the global
attractor is stochastically independent of the value $x_d$ in dimension $d$. In other words: If one looks
only on the dimension $d$, the global attractor is chosen uniformly at random from the set of all particles. 
As a consequence, after some iterations, the $d_0$'th coordinate of the velocity becomes positive for every particle,
so the attraction towards the global attractor always goes into the same direction as the velocity, while in the
remaining dimensions, the velocities may as well point away from the global attractor, meaning that the particles will
be slowed down by the force of attraction. An overview over the situation is given in Fig. \ref{PotBild}.

So, roughly speaking, most of the time the global attractor is somewhere in the middle of the different $x_d$ 
values, giving less potential increase then in dimension $d_0$ where it has a border position. That means that
the balanced situation is not stable in a sense that after the imbalance has reached a certain critical value,
it will grow unbounded.

If at some point no more improvements can be made in dimension $d_0$, the swarm is in the situation described 
above where it starts to converge while the convergence is, other than the acceleration phase, balanced. That
means after the same time the potential of every dimension is decreased by approximately the same factor, so 
dimension $d_0$ has still far more potential than any other dimension and the swarm stays blind for possible 
improvements in dimensions other than $d_0$.

To supplement the results about the behavior of the PSO in that ``artificial'' setting, we ran it on two well-known
benchmark functions to show that the problems described above really occurs on actual instances.
Since the described scenario may happen with positive but, depending on the situation, small probability, we choose the number of particles $N$ compared to the number of dimensions $D$ small in order to be able to view the phenomenon in a preferably pure condition. Table \ref{resultsclassicS} lists our results on the sphere function with optimal solution $z^* = (0,...,0)$, where we distributed the particles randomly over $[-100; 100]^D$ and the velocities over $[-50,50]^D$, and another common benchmark, the Rosenbrock function with optimal solution $z^*=(1,...,1)$ (found in \cite{Rosenbrock60}), where the initial population was randomly distributed over $[-5; 10]^D$ and the initial velocity over $[-2.5,5]^D$. The results obtained on this function are stated in Table \ref{resultsclassicR}. We repeated each experiment $1000$ times and calculated the means. Additionally we calculated for each repetition the dimension with the minimal and the one with the maximal value for the potential $\Phi$ after the last iteration (see columns $\Phi$) and the difference between the global attractor and the optimal solution in the dimension with the lowest resp. highest remaining potential. One can see that the dimension with the highest value for $\Phi$ usually is much closer to its optimal value than the dimension with the lower value. In particular, in the $2$-dimensional case the potential became $0$ in one dimension preventing the swarm from any movement in this direction and consequently from finding the minimum. 

%
\begin{table}[htb]\centering
\caption{\label{resultsclassicS}Sphere-function}
\smallskip
\begin{scriptsize}
\begin{threeparttable}
\begin{tabular}{|c||c|c|c|c|}\hline
$D$      			      &               & $4$                &  $60$  		&  $150$ 		\\\hline
$N$      			      &               & $2$                &  $10$  		&  $20$  		\\\hline
$t_{\max}$			      &               & $10000$            &$100000$		&$100000$		\\\hline
Value    			      &               & $51.04$            & $12.18$		& $11.97$		\\\hline
\multirow{2}{*}{min. $\En$}   	      &$\En$          & $0$\tnote{*}       &$5.84\cdot 10^{-62}$& $4.15\cdot 10^{-37}$	\\\cline{2-5}
			              & dist. opt.    & $1.58$             & $1.32$   		&  $1.30$      		\\\hline
\multirow{2}{*}{max. $\En$}   	      &$\En$          & $3.75\cdot10^{-8}$ & $7.53\cdot 10^{-8}$& $1.59\cdot 10^{-7}$	\\\cline{2-5}
			              & dist. opt.    & $1.16\cdot10^{-8}$ & $1.91\cdot 10^{-9}$& $2.28\cdot 10^{-9}$	\\\hline
\end{tabular}
\begin{tablenotes}\footnotesize 
\item[*] Due to double precision.
\end{tablenotes}
\end{threeparttable}
\end{scriptsize}
\end{table}
\begin{table}[htb]\centering
\caption{\label{resultsclassicR}Rosenbrock-function}
\smallskip
\begin{scriptsize}
\begin{threeparttable}
\begin{tabular}{|c||c|c|c|c|}\hline
$D$      			      &               & $4$        		& $60$       	    &  $150$   		\\\hline
$N$      			      &               & $2$        		& $10$       	    &  $20$    		\\\hline
$t_{\max}$			      &               & $10000$    		&$100000$    	    & $100000$  	\\\hline
Value    			      &               & $126.54$   		& $34.57$    	    & $28.88$   	\\\hline
\multirow{2}{*}{min. $\En$}   	      &$\En$          &$0$\tnote{*}		&$6.27\cdot 10^{-5}$& $2.32\cdot10^{-4}$\\\cline{2-5}
			              & dist. opt.    & $1.1075$   		& $0.37$     	    & $0.12$     	\\\hline
\multirow{2}{*}{max. $\En$}	      &$\En$          & $4.72\cdot 10^{-5}$ 	& $0.93$     	    & $8.07$      	\\\cline{2-5}
			              & dist. opt.    & $2.59$	        	& $0.11$       	    & $0.06$    	\\\hline
\end{tabular}
\begin{tablenotes}\footnotesize 
\item[*] Due to double precision.
\end{tablenotes}
\end{threeparttable}
\end{scriptsize}
\end{table}
We calculated the standard deviation for each unit and obtained values that were of the same order but usually higher than the mean values. The reason for this high deviations is that the examined phenomenon occurs randomly and therefore one cannot predict the potential level of the swarm when it occurs, i.\,e., the level of imbalance at the point when the swarm starts to converge is unpredictable.

Now that we know that the swarm tends to optimize functions dimensionwise, it is interesting to see what happens
if we try it on a function that is in some area increasing in every dimension but still decreasing in some direction not parallel to one of the axes.

Fix some $b>1$ and define the $D$-dimensional function $f$ as follows:
$$
f(\vec x) = \begin{cases} \sum_{i=1}^n x_i^2, \quad\exists i,j: x_i\ge b\cdot x_j \vee x_j \ge b\cdot x_i\\
		       \dfrac{\sum_{i=1}^n x_i^2}{b-1}\cdot\left(2\operatorname{max}_{i\ne j}\left\{\dfrac{x_i}{x_j}\right\}-b-1\right), \quad\!\text{otherwise}\end{cases}
$$

\begin{figure}[htb]
\centering
\includegraphics[width=10cm,height=10cm]{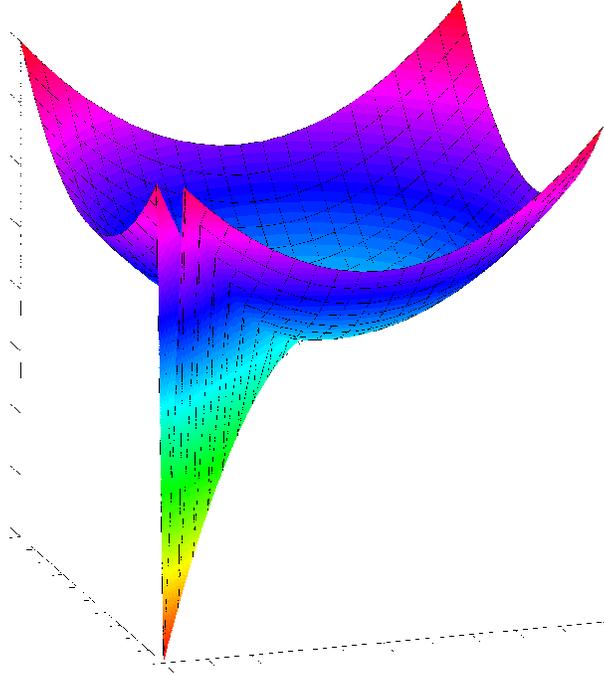}
\caption{\label{pizza}Continuous, not differentiable function~$f$}
\end{figure}

In Fig. \ref{pizza} one can see a plot of $f$ for $b=1.1$ and $D=2$. For $y$ not between $x/b$ and $x\cdot b$, this function behaves like the well-known sphere-function, leading the particles close to the origin. For $x=y$, $f(x,y)=-2\cdot x^2$ and from $y=x/b$ ($y=x\cdot b$) to $y=x$, the function falls into a valley. One can easily verify that this function is, though not differentiable, at least continuous. One would want the particles to be able to move through the valley.

\begin{figure*}[htb]
\centering
\subfloat[\label{pizzaschlecht}$f$ with $b=1.1$, $10$ particles]
{\includegraphics[width=5cm,height=5cm]{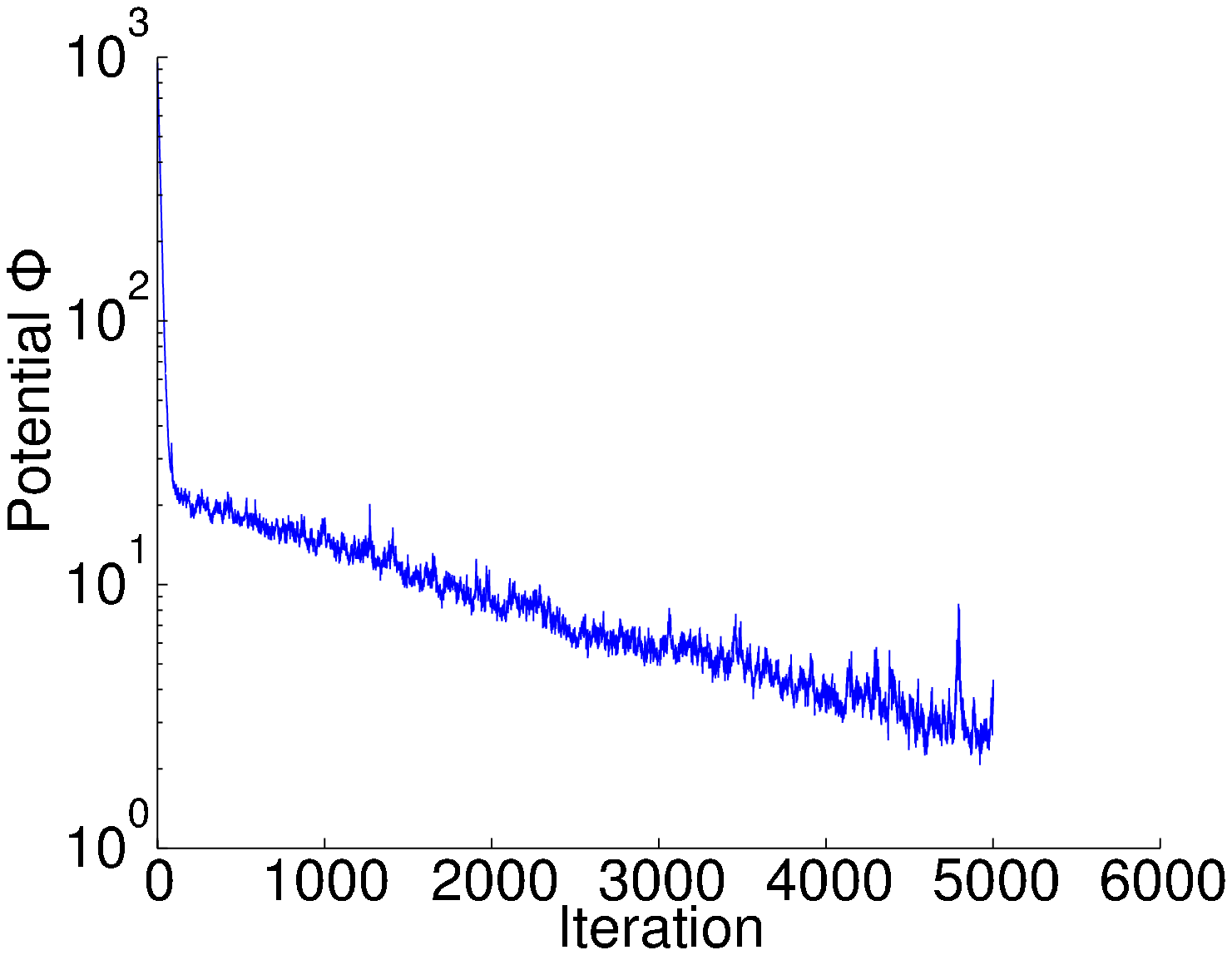}}
\subfloat[\label{pizzagut}$f$ with $b=1.1$, $50$ particles]
{\qquad\includegraphics[width=5cm,height=5cm]{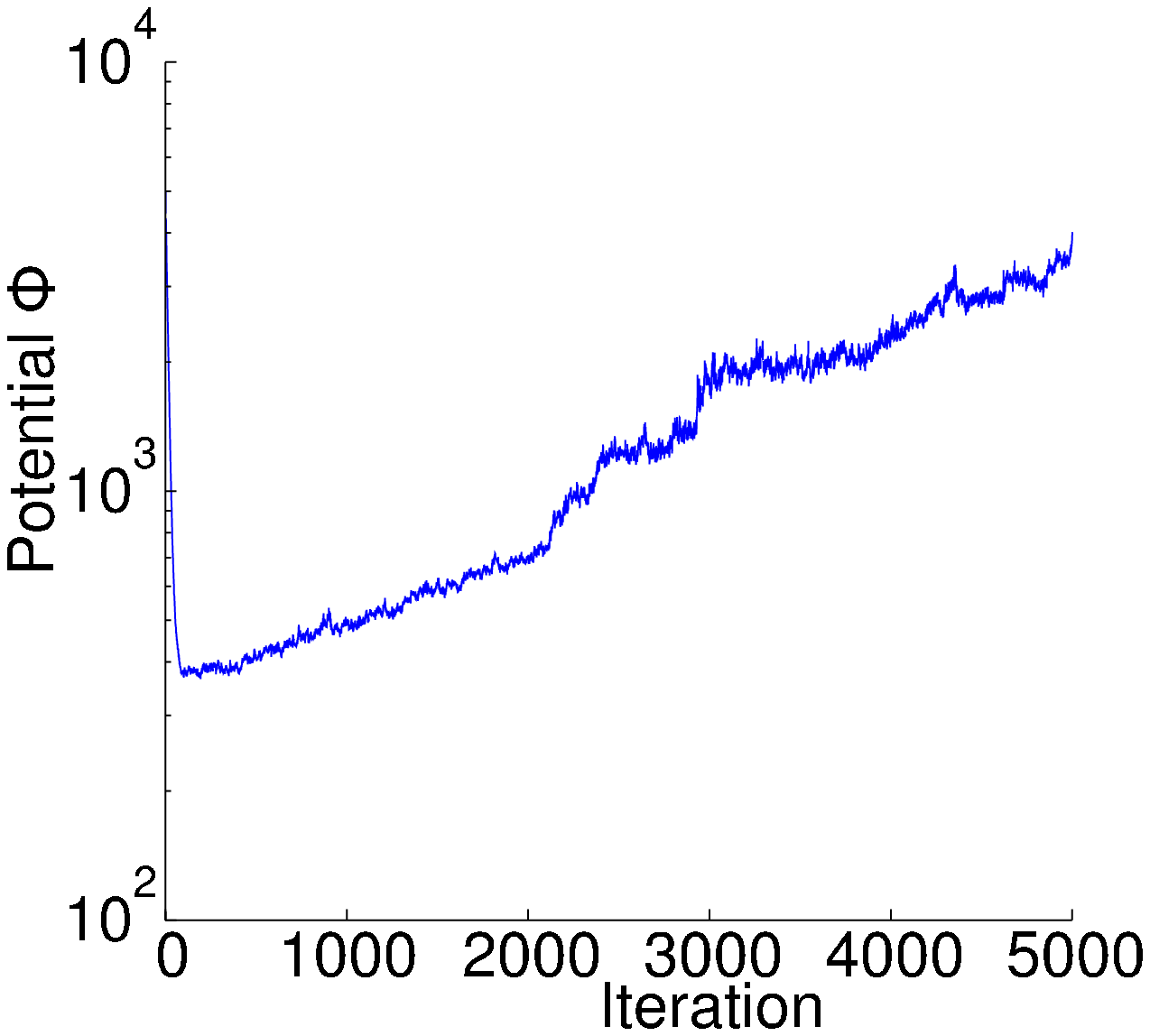}}
\subfloat[\label{pizzarot}$f_{\text{rot}}$ with $b=1.1$. $10$ particles]
{\qquad\includegraphics[width=5cm,height=5cm]{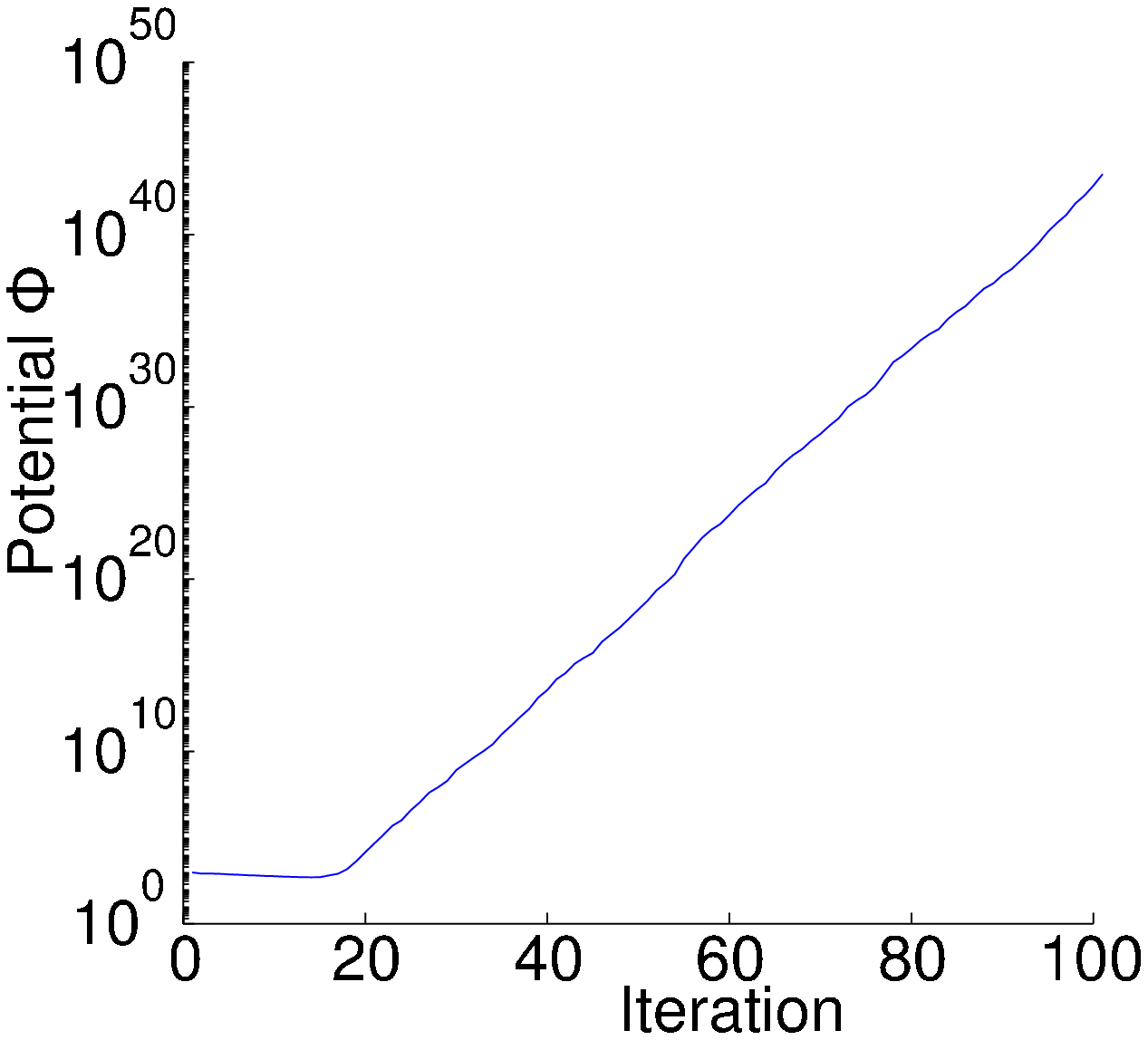}}
\caption{Behavior of the particles on functions $f$ and $f_{\text{rot}}$}
\label{pizzaruns}
\end{figure*}

As in our previous experiment, we set $\chi = 0.729$, $c_1 = c_2 = 1.49$, $N = 10$ and initialized the particles
uniformly at random over $[-100; 100]^D$ (except for the first particle that was initialized at $(1,...,1)$ such that the
swarm could see the direction where the improvements are possible) and the velocities over $[-50; 50]^D$, with the value $D=3$. 
We let the swarm do $1000$ runs with $5000$ iterations each. The potential of the dimension with the highest potential after 
the last iteration was determined and the mean and standard deviation of the respective dimensions were calculated over the 
$1000$ repetitions. This was done for two different swarm sizes, namely $N=10$ and $N=50$. 
We repeated the experiment with $10$ particles and only $100$ iterations, using the function $f_{rot}$, which is obtained by first rotating the input vector and then applying $f$ such that the valley 
now leads the particles along the $x_1$-axis. Formally speaking, the rotation maps the vector $(\sqrt{N},0,\dots,0)$ to 
$(1,1,\dots,1)$ and keeps every vector that is orthogonal to this two invariant.
The results of the three experiments can be seen in Fig. \ref{pizzaruns}. In all three cases, for about the first $20$ iterations, 
the swarm behaves like on the sphere function and reduces its potential. Then, it discoveres the valley and tries to move through 
it. However, in the unrotated case with $10$ particles (Fig. \ref{pizzaschlecht}), the swarm fails to accelerate and instead, it
converges towards a non-optimal point. With much more effort, the swarm consisting of $50$ particles (Fig. \ref{pizzagut}) is able to accelerate,
but the acceleration rate and therefore the speed are comparatively poor. Finally, Fig. \ref{pizzarot}
shows how the swarm handles the rotated version much better than the original function $f$ before. Here, after only $100$ iterations,
the potential increased to a value of about $10^{45}$. The reason for this large difference between the behavior on $f$ and on 
$f_{rot}$ is the capability of the swarm to favor one direction only if this direction is parallel to one of the axes. 

In particular, this experiment shows 
that PSO is not invariant under rotations of the search space. 

\section{Modified PSO}
\label{sec:run}

In the previous section, we have seen that the particle swarm might get stuck if its potential is too high in dimensions that are already optimized and too low in dimensions that could still be improved. Then the global attractor stagnates and the swarm starts to converge. Since the convergence happens in a symmetric manner along the different dimensions, the imbalance is maintained. A small and simple modification of the PSO algorithm avoids that problem by enabling the swarm to rebalance the potentials in the different dimensions:

\begin{defi}[Modified PSO]\label{mPSO}
For some arbitrary small but fixed $\delta>0$, we define the modified PSO via the same equations as the classic PSO in
Def.~\ref{cPSO}, only modifying the velocity update in line \ref{velup} of Algorithm \ref{alg:clPSO} to
$$
V_d^n :=
\begin{cases}
(2\cdot \operatorname{rand}()-1)\cdot \delta,\quad
	\hspace*{0.1cm}
	\text{if }\forall\, d'\in\{1,...,D\}:\left|V_{d'}^n\right|+\left|G_{d'}-X_{d'}^n\right| < \delta,\\
\vphantom{.}\\
\chi\cdot V_d^n + c_1\cdot \operatorname{rand}() \cdot (L_d^n-X_d^n)
	+ c_2\cdot \operatorname{rand}() \cdot (G_d-X_d^n), \quad
	\hspace*{0.1cm}\text{ otherwise}.
\end{cases}
$$

\end{defi}

In words: As soon as the sum of the velocity and the distance between the position and the global
attractor of a particle are below the bound of $\delta$ in every single dimension, the updated
velocity of this particular particle is drawn u.\,a.\,r. from the interval $[-\delta,\delta]$. Note
the similarity between this condition and the definition of the potential.
Indeed, we could have used the 
condition $\En_d<\delta$ (with some fixed $a$)
instead, but we decided to keep the modification as simple 
and independent from the terms occurring in the analysis as possible.
Now the potential can no longer converge to $0$ while staying unbalanced because if it decreases below a certain bound,
we randomly assign a value to the velocity which on expectation has an absolute value of $\delta/2$.
If a velocity is assigned that way, we call the step forced.

This modified PSO is similar to the Noisy PSO proposed by Lehre and Witt in \cite{LW:11} where they 
generally add a random perturbation drawn u.\,a.\,r. from $[-\delta/2,\delta/2]$ for some small $\delta$ and prove that their
swarm is able to find a local optimum. However, their analysis is restricted to
one specific $1$-dimensional fitness function.

The modification does not prevent the PSO from emerging imbalance between the potentials of different dimensions. But the imbalanced convergence phenomenon described above is no longer possible. When the global attractor of the modified PSO gets stuck and the potential decreases, there will be a point when both the velocities and the distances to the global attractor in every dimension get lower than $\delta$. From that moment on, the relationship between the different dimensions gets balanced by the forced steps which on expectation give every dimension the same amount of potential. So, the potentials of the different dimensions are made equal.

We repeated the experiment from the previous section in the same setting as before, but using the modified PSO. The results can be seen in Table \ref{resultscomp}. It turns out that the modified PSO algorithm actually leads to a better solution than the unmodified one.
\begin{table}[htb]\centering
\caption{\label{resultscomp}Comparison between the classic and the modified PSO algorithm}
\smallskip
\begin{scriptsize}
\begin{threeparttable}
\begin{tabular}{|c|c|c|c|c||c|c|c|c|c|}\hline
Function & $D$ & $N$ &$t_{\max}$ & $\delta$    & Value  \\\hline
Sphere   & $4$ & $2$ & $10000$    & $10^{-12}$ & $43.34$  \\\hline
Sphere   & $4$ & $2$ & $10000$    & -           &  $51.04$  \\\hline
Sphere   & $60$ & $10$ & $100000$    & $10^{-12}$ & $4.07$   \\\hline
Sphere   & $60$ & $10$ & $100000$    & -           & $12.18$ \\\hline
Sphere   & $150$ & $20$ & $100000$    & $10^{-12}$ & $6.41$ \\\hline
Sphere   & $150$ & $20$ & $100000$    & -          & $11.97$ \\\hline
Rosenbrock& $4$ & $2$ & $10000$    & $10^{-7}$  & $8.80$  \\\hline
Rosenbrock& $4$ & $2$ & $10000$    & -           & $126.54$   \\\hline
Rosenbrock& $60$ & $10$ & $100000$    & $10^{-7}$   & $2.02$ \\\hline
Rosenbrock& $60$ & $10$ & $100000$    & -           & $34.57$  \\\hline
Rosenbrock& $150$ & $20$ & $100000$    & $10^{-3}$   & $2.25$  \\\hline
Rosenbrock& $150$ & $20$ & $100000$    & -           & $28.88$  \\\hline
\end{tabular}
\begin{tablenotes}\footnotesize 
\item[1] Due to double precision.
\end{tablenotes}
\end{threeparttable}
\end{scriptsize}
\end{table}

 We also calculated the standard deviation for each unit and obtained values that were of the same order but usually higher than the mean values. The reason for this high deviations is that the examined phenomenon occurs randomly and therefore one cannot predict the potential level of the swarm when it occurs.
 
\begin{figure}
\centering
\includegraphics[width=8cm,height=8cm]{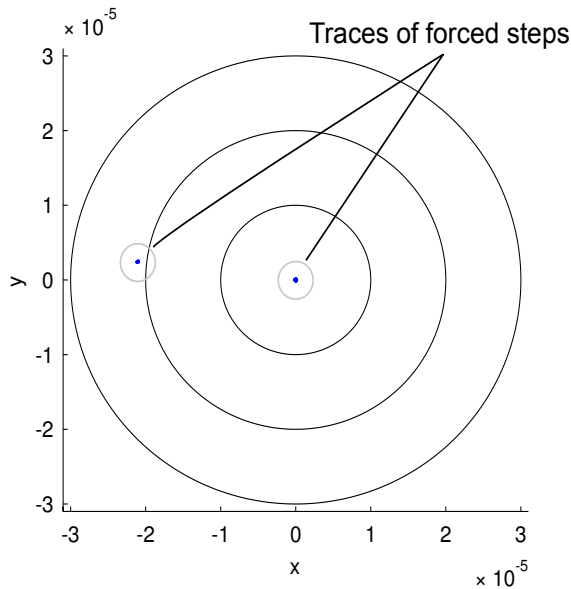}
\caption{Behavior of the modified PSO on the sphere function}\label{sphererun}
\end{figure}

To make sure that the modification does not fully take over, we plotted the forced points with $\delta = 10^{-7}$ and the $2$-dimensional sphere function as objective function in Fig. \ref{sphererun}. As can be seen in the figure, the particles get forced near $(-2\cdot 10^{-5},0)$ but their movement does not stay forced. Instead, the swarm becomes running again until the particles approached the optimum at $(0,0)$. This implies that for sufficiently smooth functions, the modification does not take over, replacing the PSO by some random search routine. Instead, the modification just helps to overcome ``corners''. As soon as there is a direction parallel to an axis with decreasing function value, the swarm becomes ``running'' again and the unmodified movement equations apply.

\section{Conclusion} 

This paper focuses on the behavior of a particle swarm to find good regions in the search space. We found out that the potentials of the different dimensions are most likely to become unbalanced and that this imbalance possibly causes the particle swarm to get stuck at non-optimal search points. A suggestion to modify the algorithm by randomly assigning a small velocity if the potential of a particle falls below a certain bound is suggested.  Additionally, we show that the modification does not take over the swarm, it just corrects the direction before the classic movement equations are applied again. 



\bibliographystyle{alpha}
\bibliography{litarXiv}

\end{document}